%% file: mlsp_template.tex
\documentclass{article}
\usepackage{amsmath,graphicx,mlspconf,amssymb,amsfonts,amsthm}

\makeatletter
\g@addto@macro \normalsize{%
	\setlength\abovedisplayskip{4pt plus 0pt minus 0pt}%
	\setlength\belowdisplayskip{4pt plus 0pt minus 0pt}}%
\makeatother

\usepackage[nodisplayskipstretch]{setspace}
\toappear{2024 IEEE International Workshop on Machine Learning for Signal Processing, Sept.\ 22--25, 2024, London, UK}

\title{Novel Gradient Sparsification Algorithm via Bayesian Inference}
\twoauthors{%
    Ali Bereyhi and Ben Liang
    \thanks{This work was funded in part by Ericsson Canada and by the Natural Sciences and Engineering Research Council of Canada.}
    \thanks{979-8-3503-7225-0/24/\$31.00 ©2024 IEEE}
}{%
    ECE Department \\
    University of Toronto \\
   \small {ali.bereyhi@utoronto.ca}, {liang@ece.utoronto.ca}
}{%
   Gary Boudreau and Ali Afana
}{%
Ericsson Canada\\
Ottawa, Canada\\
  \small \{gary.boudreau,ali.afana\}@ericsson.com 
}
\usepackage{bibentry}
\usepackage[nolist]{acronym}
\usepackage{algorithm}
\usepackage{algorithmic}
\usepackage{dsfont}
\usepackage{bbm}

\usepackage{xspace}
\newcommand{\tpK}{\textsc{Top}-$k$\xspace}
\newcommand{\rgtpK}{\textsc{RegTop}-$k$\xspace}
\usepackage{paralist}
\include{commands}

\let\oldpar\paragraph
\renewcommand{\paragraph}[1]{\vspace{-4mm}\oldpar*{#1}}

\begin{document}

\maketitle

	\begin{acronym}
	\acro{mimo}[MIMO]{multiple-input multiple-output}
	\acro{simo}[SIMO]{single-input multiple-output}
	\acro{csi}[CSI]{channel state information}
	\acro{awgn}[AWGN]{additive white Gaussian noise}
	\acro{iid}[i.i.d.]{independent and identically distributed}
	\acro{uts}[UTs]{user terminals}
	\acro{ps}[PS]{parameter server}
	\acro{irs}[IRS]{intelligent reflecting surface}
	\acro{tas}[TAS]{transmit antenna selection}
	\acro{glse}[GLSE]{generalized least square error}
	\acro{rhs}[r.h.s.]{right hand side}
	\acro{lhs}[l.h.s.]{left hand side}
	\acro{wrt}[w.r.t.]{with respect to}
	\acro{rs}[RS]{replica symmetry}
	\acro{mac}[MAC]{multiple access channel}
	\acro{np}[NP]{non-deterministic polynomial-time}
	\acro{papr}[PAPR]{peak-to-average power ratio}
	\acro{rzf}[RZF]{regularized zero forcing}
	\acro{snr}[SNR]{signal-to-noise ratio}
	\acro{sinr}[SINR]{signal-to-interference-and-noise ratio}
	\acro{svd}[SVD]{singular value decomposition}
	\acro{mf}[MF]{matched filtering}
	\acro{gamp}[GAMP]{generalized AMP}
	\acro{amp}[AMP]{approximate message passing}
	\acro{vamp}[VAMP]{vector AMP}
	\acro{map}[MAP]{maximum-a-posterior}
	\acro{ml}[ML]{maximum likelihood}
	\acro{mse}[MSE]{mean squared error}
	\acro{mmse}[MMSE]{minimum mean squared error}
	\acro{ap}[AP]{average power}
	\acro{ldgm}[LDGM]{low density generator matrix}
	\acro{tdd}[TDD]{time division duplexing}
	\acro{rss}[RSS]{residual sum of squares}
	\acro{rls}[RLS]{regularized least-squares}
	\acro{ls}[LS]{least-squares}
	\acro{erp}[ERP]{encryption redundancy parameter}
	\acro{zf}[ZF]{zero forcing}
	\acro{ta}[TA]{transmit-array}
	\acro{ofdm}[OFDM]{orthogonal frequency division multiplexing}
	\acro{dc}[DC]{difference of convex}
	\acro{bcd}[BCD]{block coordinate descent}
	\acro{mm}[MM]{majorization-maximization}
	\acro{bs}[BS]{base-station}
	\acro{aircomp}[AirComp]{over-the-air-computation}
	\acro{ULA}[ULA]{uniform linear array}
	\acro{fl}[FL]{federated learning}
	\acro{ota-fl}[OTA-FL]{over-the-air federated learning}
	\acro{los}[LoS]{line-of-sight}
	\acro{nlos}[NLoS]{non-line-of-sight}
	\acro{aoa}[AoA]{angle of arrival}
	\acro{nn}[NN]{neural network}
	\acro{cnn}[CNN]{convolutional neural network}
	\acro{dnn}[DNN]{deep neural network}
	\acro{sgd}[SGD]{stochastic gradient descent}
	\acro{aircomp}[AirComp]{over-the-air computation}
	\acro{lmi}[LMI]{linear matrix inequalities}
	\acro{qcqp}[QCQP]{quadratically constrained quadratic programming}
	\acro{lln}[LLN]{law of large numbers}
	\acro{clt}[CLT]{central limit theorem}
	\acro{gs}[GS]{gradient sparsification}
	\acro{ota-f}[OTA-F]{over-the-air fair}
	\acro{bc}[BC]{broadcast channel}
	\acro{pdf}[PDF]{probability density function}
	\acro{fedavg}[FedAvg]{federated averaging}
\end{acronym}

\begin{abstract}
	Error accumulation is an essential component of the \tpK sparsification method in distributed gradient descent. It implicitly scales the learning rate and prevents the slow-down of lateral movement, but it can also deteriorate convergence. This paper proposes a novel sparsification algorithm called \textit{regularized} \tpK (\textsc{RegTop-}$k$) that controls the learning rate scaling of error accumulation. The algorithm is developed by looking at the gradient sparsification as an inference problem and determining a Bayesian optimal sparsification mask via maximum-a-posteriori estimation. It utilizes past aggregated gradients to evaluate posterior statistics, based on which it prioritizes the local gradient entries. Numerical experiments with ResNet-18 on CIFAR-10 show that at $0.1\%$ sparsification, \rgtpK achieves about $8\%$ higher accuracy than standard \tpK. 
\end{abstract}
\section{Introduction}
\label{sec:intro}
Consider a distributed \ac{sgd} setting \cite{alistarh2018convergence}, where $N$ workers compute gradients and share them with a server to estimate a \textit{global} gradient. In iteration $t$, worker $n$ starts from a common model $\btheta^t\in\setR^J$ and computes its \textit{local} gradient using a local surrogate loss function determined by averaging the loss over a stochastically-selected mini-batch. Let $\bg_n^{t}\in\setR^J$ denote the gradient computed by worker $n$. After receiving $\bg_1^{t}, \ldots, \bg_N^t$, the server estimates the global gradient $\bg^t$ by weighted averaging, i.e., %
\begin{align}
	\bg^{t} =   \sum_{n=1}^N \omega_n  \bg_n^{t} %
\end{align}
for some non-negative weights $\omega_n$. The estimated gradient is used to update the common model as $\btheta^{t+1} = \btheta^t - \eta^t \bg^{t}$ with some learning rate $\eta^t$. 

With realistic models, the scale of communication in such settings can be prohibitive, e.g., for ResNet-110 $J\approx1.7 \times 10^6$ \cite{he2016deep}. Assuming $1000$ mini-batches at each worker, the network exchanges $1.7 \times 10^9$ symbols per epoch for each worker. A classical solution to this issue is \textit{gradient sparsification} \cite{alistarh2018convergence,strom2015scalable}: in each iteration, the workers only send \textit{important gradient entries} along with their indices. The term \textit{sparsification} refers to the fact that this approach 
can be interpreted as if the sever approximates local gradients with their \textit{sparsified version}. Typical order of sparsity in practice is less than $1\%$, i.e., fewer than $0.01J$ entries are sent by each worker \cite{strom2015scalable,dryden2016communication,aji2017sparse,lindeep,sahu2021rethinking}. 

The key point in gradient sparsification is the design of a mechanism which can efficiently find \textit{important gradient entries}. This can be challenging, as each worker has no explicit information about the gradients of other workers, and hence decides \textit{locally}. In this work, we address this challenge by developing a new algorithm called \rgtpK, which extracts \textit{global} information from earlier iterations. \rgtpK can be seen as the classical \tpK with regularization that controls the \textit{learning rate scaling property} of error accumulation. We illustrate this property next.

\subsection{Error Accumulation and Learning Rate Scaling}
The standard approach for sparsification is \tpK, which selects the $k$ largest \textit{accumulated} gradient entries in each iteration. 
In iteration $t$, worker $n$ computes its local gradient $\bg_n^t$ and the accumulated gradient as $\baa_n^t = \beps_n^{t} + \bg_n^t$, where $\beps_n^{t}$ is the \textit{sparsification error} from the previous iteration. It then selects the $k$ entries of $\baa_n^t$ with the largest amplitude which results in the sparsified gradient $\hat{\bg}_n^t \in\setR^J$.  The sparsification error is then updated as $\beps_n^{t+1} = \baa_n^t - \hat{\bg}_n^t$.

As one may notice, beyond naive selection of the $k$ largest gradient entries, a key procedure in \tpK is \textit{error accumulation}, i.e., the computation of sparsification error $\beps_n^{t} $. 
This way, the initially unselected entries get the chance of being selected after their errors become large enough. After such an entry is eventually selected, \ac{sgd} moves a large step in the entry's direction, whose length is proportional to the gradient of that entry  \textit{accumulated} in previous iterations. This behavior is known as \textit{learning rate scaling} \cite{lin2018deep}. Though this scaling is fairly effective for smooth losses, for other losses it can result in either alternation around the optimum or divergence.  This behavior is best understood through a toy example that is given in the sequel.

\subsection{A Motivational Example}
Consider logistic regression with $J=2$, where $N=2$ workers employ distributed gradient descent to minimize the cross-entropy. Let worker 1 and 2 have single data-points $\brc{\bxx_1, 1}$ and $\brc{\bxx_2, 1}$, respectively, where $\bxx_1 = \dbc{100,1},$ and $\bxx_2 = \dbc{-100,1}$. The workers agree on a model with weight vector $\btheta = \dbc{\theta_1, \theta_2}$  and zero bias. The loss of worker $n\in\set{1,2}$ in this case is 
	$F_n \brc{\btheta} =  \log \brc{1+ \exp\set{- \inner{\btheta;\bxx_n} }}$, 
and the empirical risk used for distributed training is  $F \brc{\btheta} =\brc{F_1 \brc{\btheta} + F_2 \brc{\btheta}}/2$. Worker $n$ shares its gradient 
\begin{align}
	\bg_n = -\frac{  \exp\set{- \inner{\btheta;\bxx_n} }  \bxx_n }{ 1+ \exp\set{- \inner{\btheta;\bxx_n} } },
\end{align}
with the server who computes $\bg = 0.5\brc{\bg_1 + \bg_2} $. 

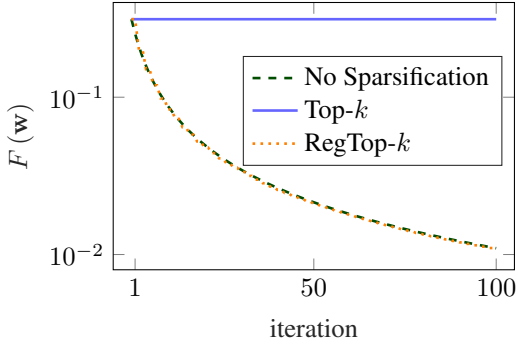
\begin{figure}
	\begin{center}
		\input{Figures/Example_1.tex}
	\end{center}
	\vspace{-3.5mm}
	\caption{Example of large learning rate scaling in \tpK.\vspace{-3.5mm}}
\label{fig:toy}
\end{figure}

Figure~\ref{fig:toy} shows training loss against iterations for learning rate $\eta = 0.9$ and $\btheta^{0} = \dbc{0, 1}$ for both \textsc{Top-}$1$ sparsification and non-sparsified cases. It is observed that \textsc{Top-}$1$ is not able to reduce the empirical risk even after $100$ iterations. This behavior can be explained as follows: at $\btheta^{0}$, the gradients are $\bg_1 = 0.736 \dbc{- 100,  1}$ and $\bg_2 = 0.736 \dbc{ 100,  1}$, and \textsc{Top-}$1$ selects the first entry at both workers. One can however see that despite their significantly larger amplitudes, the first entries do not contribute in the training, as they cancel out after averaging. With \textsc{Top-}$1$, the aggregated sparsified gradient remains zero, and hence the distributed gradient descent remains at $\btheta^0$ for several iterations. It starts to move from the initial point only after a large number of iterations when the accumulated sparsification error at the second entry starts to surpass the first entry in amplitude. For comparison, we further show the results for \textsc{RegTop-}$1$: our proposed algorithm tracks non-sparsified training consistently.

One can readily extend this toy-example to a setting where learning rate scaling hinders the convergence. 
For instance, let the loss of worker $n$ be $\tilde{F}_n \brc{\btheta} = F_n \brc{\btheta} + G\brc{\theta_2}$, for some $G\brc{\theta_2}$ whose derivative at $\theta_2=1$ is $1$. Starting from $\btheta^{0} = \dbc{0, 1}$, \textsc{Top-}$1$ aggregates zero gradients in the first $50$ iterations and moves from $\btheta^{0}$ only at $t = 51$ when the accumulation error at both workers read $\beps_n^{t} = \dbc{0,100}$. At this iteration, the workers send their sparsified accumulated gradients $\hat{\bg}_n^{t} = \dbc{0,100}$, which leads to ${\bg}^t = \dbc{0,100}$. Comparing this gradient with the non-sparsified aggregation in the first iteration, one can see that \textsc{Top-}$1$ scales the learning rate with factor $50$. Depending on $G\brc{\theta_2}$, this large scaling of the learning rate can deteriorate the convergence of the optimizer. 

\subsection{Related Work}
\label{sec:2}
Several lines of work have extended distributed \ac{sgd} with \tpK: the study in \cite{chen2018adacomp} proposes an adaptive sparsification technique based on \tpK aiming to reduce the computational complexity. The authors of \cite{lin2018deep} develop the \textit{deep gradient compression scheme} that incorporates the ideas of momentum correction into \tpK. Online adaptation of \tpK with the goal of having minimum training time is discussed in \cite{han2020adaptive}. In \cite{wang2018atomo}, the \tpK extension \textit{Atomo} is introduced, which sparsifies the gradients in an arbitrary atomic decomposition space. 
Layer-wise gradient sparsification for \acp{dnn} is further investigated in \cite{zhang2022mipd}. 
 In \cite{chen2020scalecom}, the authors propose a scalable sparsified gradient compression that uses the similarity of local gradients to enhance the scalability of \tpK. 
 The above lines of work substantially differ from our study as \textit{they mainly focus on adapting the} 
 \textit{classical \tpK to a wider range of distributed settings}, e.g., various optimizers. Unlike these lines of work, our study aims to develop a new sparsification scheme that \textit{controls  the learning rate scaling}.

Developing sparsification algorithms based on model statistics has been recently investigated in \cite{sahu2021rethinking} and \cite{m2021efficient}. In \cite{sahu2021rethinking}, the authors revisit \tpK sparsification and give an alternative interpretation as the optimal sparsifier for per-iteration communication budget. The notion of optimality is then extended to the entire training leading to a new \tpK based sparsification scheme. The study in \cite{m2021efficient} proposes a statistical approach for gradient sparsification by treating local gradients as random variables distributed with empirically-validated sparsity-inducing distributions. 
It is worth mentioning that, these studies do \textit{not} propose any mechanism for controlling the learning rate scaling and mainly focus on extending \tpK sparsification under a new set of design constraints. To our best knowledge, \textit{our work is the first study that develops a sparsification algorithm based on error accumulation that controls learning rate scaling.} 

\subsection{Contributions}
In this work, we develop a Bayesian framework for gradient sparsification. Unlike earlier studies, we focus on the learning rate scaling of \tpK and propose a regularization technique to control this property. 
In a nutshell, our main contributions are as follows:
\begin{inparaenum}
\item[(1)] We formulate gradient sparsification as an inference problem. Invoking this formulation, we represent the Bayesian optimal sparsifier as a \ac{map} estimator. 
\item[(2)] We construct a prior belief on the gradient entries by interpreting \tpK as a \textit{mismatched} \ac{map} sparsifier with postulated uniform likelihood. 
We call this prior distribution the \tpK prior belief.
\item[(3)] Using the \tpK prior belief, we determine the optimal sparsification mask and show that it is a regularized form of the \tpK sparsifier.
\end{inparaenum} 
We validate our derivations through numerical experiments, which suggest that while \tpK oscillates at a fixed optimality gap, \rgtpK can converge to the global optimum with significantly sparser local gradients.

\paragraph{Notation}
Vectors are shown in bold, e.g., $\bxx$. The $j$-th entry of vector $\bxx$ is represented as $x\id{j}{}$. Entry-wise multiplication and division are denoted by $\odot$ and $\oslash$, respectively. The inner product of $\bxx$ and $\byy$ is represented by $\inner{\bxx;\byy}$. The $\ell_p$-norm of $\bxx$ is denoted by $\norm{\bxx}_p$. For an integer $N$, the set $\set{1,\ldots,N}$ is abbreviated as $\dbc{N}$. We denote the cardinality of set $\setS$ by $\abs{\setS}$.

\section{Preliminaries}
We consider the distributed setting described in Section~\ref{sec:intro}. Let $\setD_n = \set{ \bxx_{n,i} \; \text{ for } \; i \in \dbc{D_n}  }$ denote the training batch of size $D_n$ at worker $n\in\dbc{N}$ whose entries are sampled \ac{iid} from $p\brc{\bxx}$. The distributed training in this setting is formulated as
\begin{align}
	\min_{\btheta\in \setR^J}   \sum_{n=1}^N \omega_n F_n(\btheta) \label{eq:main}
\end{align}
for some $\omega_n$ proportional to $D_n$, where $F_n(\btheta)$ is the empirical loss computed by worker $n$, i.e.,
\begin{align}
	F_n\brc{\btheta} = \frac{1}{D_n} \sum_{i=1}^{D_n} f(\btheta\vert \bxx_{n,i})
\end{align}
for some loss function $f\brc{\btheta\vert \bxx}$. The goal is to solve this optimization in a distributed fashion with minimal communication overhead. We consider a gradient-based optimizer, where the server in iteration $t$ is interested in computing $\bg^{t} =   \sum_{n} \omega_n  \bg_n^{t}$ with  $\bg_n^{t}$ denoting the gradient of $F_n(\btheta)$ at $\btheta^t$. 

\paragraph{\tpK Sparsification}
With gradient sparsification, worker $n$ sends the sparsified gradients $\hat{\bg}_n^t =\bss_n^t \odot \baa_n^t$ to the sever, where $\bss_n^t\in\set{0,1}^J$ is the sparsification mask and $\baa_n^t$ denotes the accumulated gradient computed by adding the local gradient $\bg_n^t$ to the sparsification error $\beps_n^{t}$ as defined in Section \ref{sec:intro}. The \tpK mask is determined by selecting the $k$ largest entries of $\baa_n^t$, i.e., $\bss_n^t = \Top{k}{\baa_n^t}$, where the \textit{top $k$ selector} $\Top{k}{\cdot}$ is defined as follows: let the entries of $\bx\in\setR^J$ be sorted as $\abs{x_{i_1}} \geq \ldots  \abs{x_{i_J}}$; then, the $i$-th entry of $\Top{k}{\bx}$ is 
\begin{align}
	\Top{k}{\bx}_{\dbc{i}} = \begin{cases}
		1 &\text{if } \; i \in \set{i_1, \ldots, i_k}\\
		0 &\text{elsewhere}.
	\end{cases}
\end{align}

Upon receiving the sparsified gradients $\hat{\bg}_n^t =\bss_n^t \odot \baa_n^t$, the server estimates the global gradient as $\bg^t = \sum_{n} \omega_n \hat{\bg}_n^t$.  
Note that the communication reduction by gradient sparsification is achieved at the expense of an extra index transmission per symbol. However, the index can be losslessly represented by $\log J$ bits, so its communication can be neglected. 

\section{Bayesian Gradient Sparsification}
Though intuitive, there is no existing study that examines the optimality of \tpK sparsification. In this section, we develop a stochastic framework for gradient sparsification by interpreting it as a Bayesian inference problem. We then show that, despite its effectiveness, \tpK is not the optimal approach in this stochastic framework. We derive the \rgtpK algorithm by characterizing the optimal sparsification scheme in the Bayesian sense. Due to the lack of space, we only present the key steps in this section. Detailed derivations are skipped and left for the extended version of the paper. 

\subsection{Bayesian-Optimal Sparsification}
Let us start with a thought experiment in which a genie provides each worker information about the aggregated gradient of the other workers. For entry $j \in \dbc{J}$, worker $n$ knows in advance the weighted sum gradient of the other workers, denoted by $z\id{j}{n}^t$, which satisfies $a\id{j}{}^t = \omega_n a\id{j}{n}^t + z\id{j}{n}^t$, where $a\id{j}{}^t$ is the $j$-th aggregated entry \textit{when the workers apply no sparsification}, i.e., $a\id{j}{}^t = \sum_{n} \omega_n a\id{j}{n}^t$. Worker~$n$ decides to transmit $a\id{j}{n}^t$, only if $a\id{j}{}^t$ is within the top $k$ gradient entries. In other words, given that the workers know $z\id{j}{n}^t$, they apply \tpK directly on the average gradient. We refer to this idealized approach as \textit{global \tpK}.  
It is readily seen that global \tpK is in practice infeasible, as worker $n$ does not have access to $z\id{j}{n}^t$. 

\paragraph{Statistical Global \tpK}
Although worker $n$ does not know $z\id{j}{n}^t$, it has partial 
access to $z\id{j}{n}^\ell$ for $\ell < {t}$ through the global gradients collected in previous iterations. This can be used to estimate $z\id{j}{n}^t$, i.e., the workers use the \textit{information collected through time} to apply the global \tpK \textit{statistically}. 
In the Bayesian framework, we can formulate a \textit{statistical global \tpK} via the following principle \ac{map} problem: 

\begin{definition}[Principle MAP Problem]
	\label{def:map}
	Let $\setT_k^t$ denote the set of $k$ largest entries of $\abs{\baa^t}$. Worker $n$ determines its posterior probabilities
	\begin{align}
		P\id{j}{n}^t = \Pr\set{j \in \setT_k^t \left\vert \baa_n^t, \set{\baa_n^\ell, \bg^\ell: \ell < {t-1} } \right.}
	\end{align}
	for  $j\in\dbc{J}$, where $\bg^\ell$ is the aggregated gradient in iteration $\ell < t$ that is already known to all workers.\footnote{Note that in distributed \ac{sgd}, the server broadcasts either $\bg^t$ or $\btheta^{t+1}$ to all workers at iteration $t$. In the latter case, the workers can recover the gradient as $\bg^t = \brc{\btheta^{t+1} - \btheta^t}/\eta^t$.} Worker $n$ then selects the $k$ entries with the largest posteriors.
\end{definition}

To solve the principle \ac{map} problem, we invoke the Bayes rule. The posterior probability $P\id{j}{n}^t$ is expanded as 
\begin{subequations}
	\begin{align}
		P\id{j}{n}^t &= 
		\Pr\set{j \in \setT_k^t \left\vert  \baa_n^t, \set{\baa_n^\ell, \bg^\ell: \ell < {t} } \right.}\\
		&= 
		\mal\id{j}{n}^t
		\Pr\set{j \in \setT_k^t \left\vert \baa_n^t  \right.} \label{eq:bayes}
	\end{align}
\end{subequations}
where $\Pr\set{j \in \setT_k^t \left\vert \baa_n^t  \right.}$ is the \textit{prior belief}, i.e., the prior probability of $j \in \setT_k^t$ based on the local gradient of worker $n$, and $\mal\id{j}{n}^t$ denotes the \textit{likelihood} given by
\begin{align}
	\mal\id{j}{n}^t \propto p\brc{ \left. { \baa_n^\ell, \bg^\ell: \ell< {t} } \right\vert  j \in \setT_k^t ,  \baa_n^t}.
\end{align}
Here, we use notation $p\brc{\cdot}$ to refer to the \ac{pdf}. 

\paragraph{\tpK in Bayesian Framework}
\tpK can be seen as a mismatched form of the principle \ac{map} sparsifier under a specific prior belief. To see this, let us consider the following prior probability, to which we refer to as \textit{\tpK prior belief}.
\begin{definition}[\tpK Prior Belief]
	Given the accumulated gradient $\baa_n^t$, the probability of entry $j$ being among the top $k$ entries of $\bg^t$ is proportional to $\abs{a\id{j}{n}^t}$, i.e., 
	\begin{align}
		\Pr\set{j \in \setT_k^t \left\vert \baa_n^t  \right.} = \frac{\abs{a\id{j}{n}^t}}{\norm{\baa_n^t}_1}. \label{eq:prior}
	\end{align}
\end{definition}
\noindent With this prior belief, the \ac{map} sparsifier reduces to
\begin{align}
	\argmaxk_j P\id{j}{n}^t 
	&= \argmaxk_j
	\mal\id{j}{n}^t
	\abs{a\id{j}{n}^t} \label{eq:map-general},
\end{align}
where $\argmaxk_j x_j$ returns the $k$ largest entries of the sequence $\set{x_j: j \in \dbc{J}}$ for $J\geq k$.

Comparing \eqref{eq:map-general} with \tpK, one can readily conclude that \textit{\tpK is a \ac{map} sparsifier whose likelihood $\mal\id{j}{n}^t$ is \textit{uniform}}. This is however a mismatched assumption. In fact, \tpK simply ignores the information collected in previous iterations and infers the dominant entries solely based on the local accumulated gradients. 

\paragraph{\rgtpK Sparsification} 
It is not straightforward to determine the exact expression for likelihood $\mal\id{j}{n}^t$, since the statistical model of its forward probability problem~is~not~completely known. In the sequel, we approximate it in the large-system limit $J\to \infty$ under some simplifying assumptions. 

We start the derivations by finding an alternative expression for the posterior probability $P\id{j}{n}^t $:
\begin{proposition}
	\label{prop:1}
	The posterior $P\id{j}{n}^t $ is computed as 
	\begin{align*}
		P\id{j}{n}^t  
		= \int_{\setF_j^k }
		q_n({\baa^t}) \dif \baa^t,
	\end{align*}
	where $\setF_j^k = \set{ \bx \in \setR^J: x_j \in \argmaxk_i x_i}$ with $x_i$ denoting the $i$-th entry of $\bx$,  and $q_n({\baa^t})$ is 
	\begin{align}
		q_n({\baa^t}) = 
		p \brc{ \baa^t \left\vert \baa_n^t,  \set{\baa_n^\ell, \bg^\ell: \ell < {t} } \right.}.
	\end{align}
\end{proposition}
\begin{proof}
	The proof is given by standard marginalization and use of the fact that $\baa_n^t,  \set{\baa_n^\ell, \bg^\ell :\ell < t} \rightarrow \baa^t \rightarrow \setT_k^t$ form a Markov chain.\footnote{Note that $\setT_k^t$ is defined in Definition~\ref{def:map}.} Details are left for the extended version.
\end{proof}

Using this alternative form, 
we can write 
\begin{align}
	\mal\id{j}{n}^t \propto \frac{1 }{\abs{a\id{j}{n}^t}} \int_{\setF_j^k }
	q_n({\baa^t}) \dif \baa^t. \label{eq:Likelihood}
\end{align}
The likelihood calculation is hence reduced to the characterization of the conditional distribution $q_n\brc{\baa^t}$. 
To describe $q_n({\baa^t})$, we need to specify the stochastic model  that 
describe the relation between $\baa^t$ and ${\baa_n^t,\set{\baa_n^\ell, \bg^\ell: \ell < {t} }}$. 
Considering the gradient aggregation strategy at the server, we can write  
	$\baa^t = \omega_n \baa_n^t + \bzz_n^t$, 
where $\bzz_n^t$ denotes the vector form of $z\id{j}{n}^t$ as defined above. We now describe the time evolution of $\bzz_n^t$ via an additive model, i.e., %
$\bzz_n^t =  \bzz_n^{t-1} + \bxi_n^{t}$ %
for some \textit{innovation} $\bxi_n^{t}$, which we treat as a random variable. It is worth mentioning that the innovation describes the difference between two consecutive local gradients, and hence its distribution depends on the dataset. 

We now divide the index set $\dbc{J}$ into two subsets, namely $\setS_n^{t-1}$ and its complement, where $\setS_n^{t-1}$ denotes the support of the previous sparsification mask of worker $n$ $\bss_n^{t-1}$. We now focus on $j \in \setS_n^{t-1}$. For this set, worker $n$ has already received the aggregated gradient in the last iteration, i.e., $a\id{j}{}^t = \rmg \id{j}{}^t$. We can hence write 
\begin{align}
	z\id{j}{n}^{t-1}  
	&= \rmg\id{j}{}^{t-1} - \omega_n a\id{j}{n}^{t-1}
	= \omega_n a\id{j}{n}^{t} \Delta\id{j}{n}^{t} 
\end{align}
where we define 
	$\Delta\id{j}{n}^{t} = (\rmg\id{j}{}^{t-1} - \omega_n a\id{j}{n}^{t-1})/{\omega_n a\id{j}{n}^{t}}$,
and refer to it as the \textit{posterior distortion}. 
We now write $a\id{j}{}^t = \bar{\rmg}\id{j}{n}^t + \xi\id{j}{n}^{t}$, where 
	$\bar{\rmg}\id{j}{n}^t = \omega_n a\id{j}{n}^t (1+\Delta\id{j}{n}^{t} )$.

To proceed with the exact computation of likelihood $\mal\id{j}{n}^t$, we require an explicit expression of the distribution of $\xi\id{j}{n}^{t}$. This is however analytically infeasible, as it depends on several problem-specific factors, e.g., data distribution, learning rate, and loss function. 
We hence invoke large-deviations arguments and the method of types to asymptotically approximate the likelihood for a class of settings in which $\bxi_n^{t}$ is distributed symmetrically around zero and is fast decaying. We skip details due to lack of space. 

\begin{proposition}
	\label{prop:2}
	Let $p_j \brc{\xi}$ denote the distribution of the $j$-th entry in $\bxi_{n}^{t}$, i.e., $ \xi\id{j}{n}^{t}$. Assume that $p_j \brc{\xi}$ represents a symmetric zero-mean distribution, and that for $j\in \dbc{J}$, given a small $\delta$, there exists a small $\varepsilon$, such that 
	\begin{align}
		\int_{-\varepsilon \abs{a\id{j}{n}^t}}^{\varepsilon \abs{a\id{j}{n}^t} } p_j \brc{\xi} \dif \xi \geq 1 - \delta.
	\end{align}
	Moreover, assume that the entries of $\bxi_{n}^{t}$ are independent.~Then, as $J\to \infty$, we have
	\begin{align*}
		\int_{\setF_j^k }	q_n({\baa^t}) \dif \baa^t \propto \abs{	a\id{j}{n}^t}
		\begin{cases}
			u_\mu ({\abs{ 1 +   \Delta\id{j}{n}^{t}  } }) &j \in \setS_n^{t-1}\\
			C &j \notin \setS_n^{t-1}
		\end{cases},
	\end{align*}
	for some constant $C$, and a non-decreasing $u_\mu: \setR^+ \mapsto \setR^+$ that approximates the sign function and is tuned by a positive parameter $\mu$.
\end{proposition}
\begin{proof}
	The proof follows large-deviations arguments. We skip the details here due to lack of space and leave them for the extended version of the paper. 
\end{proof}

From Proposition~\ref{prop:2}, we can conclude that under the given assumptions, the likelihood for large $J$ is approximated by
\begin{align}
	\mal\id{j}{n}^t = u_\mu ({\abs{ 1 +   \Delta\id{j}{n}^{t}  } }). \label{asymp}
\end{align}
where, for $j \notin \setS_n^{t-1}$, we define $\Delta\id{j}{n}^{t}=Q$ for some $Q$ satisfying $u_\mu (\abs{1+Q}) = C$.  The Bayesian-optimal sparsification is hence approximated by substituting this expression into the principle \ac{map} sparsifier. 

\subsection{\rgtpK Algorithm}
The asymptotic expression \eqref{asymp} for the likelihood implies that for Bayesian-optimal sparsification, the top $k$ selector should be applied on a \textit{regularized accumulated gradient}, i.e., 
\begin{align}
\tilde{\baa}_n^t =	\baa_n^t \odot u_\mu ({\abs{ 1 +   \bdelta_n^t  } }),
\end{align}
where $\bdelta_n^t \in \setR^J$ collects the posterior distortions $\Delta\id{j}{n}^{t}$ for $j\in\dbc{J}$.
We now follow the definition of $u_\mu \brc{\cdot}$ in Proposition~\ref{prop:2} and set\footnote{Note that other choices, e.g., sigmoid function, are also valid.} $u_\mu \brc{x} =  \tanh \brc{ x / \mu}$, 
where we can treat $\mu$ as a hyperparameter. This concludes the \rgtpK algorithm, whose pseudo code is given in Algorithm~\ref{alg:RegtopK}.

\begin{figure*}
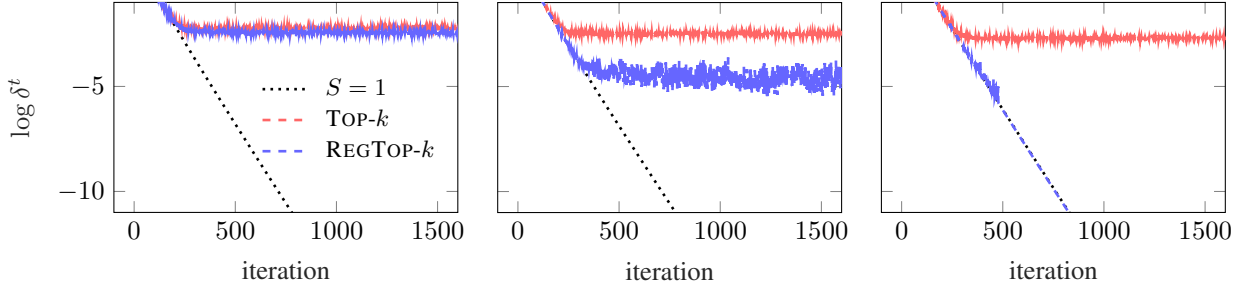

	\begin{center}
		\input{Figures/Reg_top_K_4.tex}
		\input{Figures/Reg_top_K_5.tex}
		\input{Figures/Reg_top_K_6.tex}
	\end{center}
	\vspace{-3.5mm}
	\caption{\rgtpK versus \tpK sparsification for three sparsity factors. Left: $S=0.4$; middle: $S=0.5$; right: $S=0.6$.\vspace{-3.5mm}}
	\label{fig:2}
\end{figure*}

\paragraph{Discussions on Algorithm~\ref{alg:RegtopK}}
\rgtpK starts by applying standard \tpK in the initial iteration. From $t=1$, worker $n$ after calculating its accumulated gradient $\baa_n^t$ determines the \textit{posterior distortion} $\bdelta_n^t$ for those entries that were sent in the previous iterations, i.e., $j$ for which $s\id{j}{n}^t =1$. The impact of the regularization applied by \rgtpK can be intuitively illustrated by considering the following extreme cases:
\begin{inparaenum}
	\item[(1)] As $\mu \rightarrow 0$, the regularizer converges to $1$, and hence \rgtpK reduces to standard \tpK. \tpK is hence a special case of \rgtpK with no regularization.
	\item[(2)] Assume the $j$-th local gradient entries of all workers are large in amplitude but cancel out after aggregation.\footnote{Recall the toy-example in Section~\ref{sec:intro}.} In this case, after the initial aggregation, worker $n$ determines its posterior distortion as $\Delta\id{j}{n}^t = 0- a\id{j}{n}^{t-1}/ a\id{j}{n}^{t} = -1$. This leads to the $j$-th regularized accumulated gradient entry damped to zero and prevents its selection in iteration $t$. This way the selection frequency of constructively-aggregated gradient entries with small amplitudes is increased, and hence large scaling of learning~rate~is~avoided.	
\end{inparaenum}


\begin{algorithm}[tb]
	\caption{\rgtpK Sparsification at Worker $n$}
	\label{alg:RegtopK}
	\textbf{Initialization}: Set $\beps_n^0 = \set{0}^J$ and some $Q, \mu > 0$
	
	\begin{algorithmic}[1] 
		\STATE \textbf{for} $t=0$ \textbf{do} Sparsify via \tpK
		and collect $\bg^0$ \textbf{end for}
		\FOR{$t\geq 1$}
		\STATE Determine local gradient $\bg_n^t$ at global model $\btheta^t$
		\STATE Determine \textit{accumulated gradient} as $\baa_n^t = \beps_n^{t} + \bg_n^t$
		\STATE Determine the \textit{posterior distortion} as
		\begin{align*}
			\hspace*{-2mm}\bdelta_n^t \hspace*{-.7mm}=\hspace*{-.7mm} \bss_n^{t-1}\dbc{ \brc{\bg^{t-1} - \omega_n \baa_n^{t-1} } \oslash \omega_n \baa_n^{t} } \hspace*{-.7mm}+\hspace*{-.7mm} Q \brc{1-\bss_n^{t-1}}
		\end{align*}
		\STATE Find the sparsification mask as
		\begin{align*}
			\bss_n^t = \Top{k}{ \baa_n^t \odot \tanh \brc{ \dfrac{\abs{1+ \bdelta_n^t}}{\mu}} }
		\end{align*}
		\STATE Sparsify as $\hat{\bg}_n^t =\bss_n^t \odot \baa_n^t$ and send to server
		\STATE Update the \textit{sparsification error} as $\beps_n^{t+1} = \baa_n^{t} - \hat{\bg}_n^t$
		\ENDFOR
	\end{algorithmic}
\end{algorithm}

\section{Numerical Validation}
We validate \rgtpK through numerical experiments. We start with the problem of linear regression. For this learning problem, we can track the optimal solution and hence evaluate the \textit{distance between the converging point and the global optimum}, which is a strong notion of convergence. 
We further give the results for training ResNet-18 on CIFAR-10. 
Throughout the numerical experiments, 
 \rgtpK is compared against \tpK and the distributed gradient descent without sparsification. As mentioned, the existing extensions to \tpK, e.g., \cite{sahu2021rethinking,lin2018deep,chen2018adacomp}, do not revise the derivation of sparsification mask. 
This means that with respect to impact of large learning rate scaling, these approaches perform identically to \tpK, and thus our comparison with \tpK suffices.

\subsection{Linear Regression}
We consider a distributed setting with $20$ workers that solve a distributed linear regression problem via distributed \ac{sgd}. Each worker has $500$ labeled data-points of dimension $100$. 
The workers employ the method of \ac{ls}. 
The global loss is determined by arithmetic averaging, i.e., $\omega_n = 1/N$. 
The training is performed via full-batch gradient descent. The learning rate is kept fixed at $\eta = 10^{-2}$. 

\paragraph{Synthetic Dataset Generation}
The local datasets are generated independently via a Gaussian linear model. For worker $n$, data-points are sampled independently from an \ac{iid} zero-mean and unit-variance Gaussian process. 
To label these data-points, we generate the ground truth model $\bt_n \in \setR^J$ \ac{iid} according to a Gaussian distribution with mean $u_n$ and variance $h^2$. 
The mean $u_n$ is further sampled from a Gaussian process with mean $U$ and variance $\sigma^2$. 
The labels are determined via the linear model as $y_{n,i} = \bxx_{n,i}^\trp \bt_n + \varepsilon_{n,i}$, where $\varepsilon_{n,i}$ is a zero-mean Gaussian perturbation with variance $\epsilon$. 

\paragraph{Numerical Results}
We evaluate performance by tracking the \textit{optimality gap}: in iteration $t$, we compute the difference between the globally updated model parameter, i.e., $\btheta^t$, and the global minimizer $\btheta^\star$, i.e.,  
$\delta^t = \norm{\btheta^t - \btheta^\star}$.	%
We denote the sparsification factor by $S = k/J$. 

Figure~\ref{fig:2} sketches the optimality gap $\delta^t$  in the logarithmic scale against the number of iterations for the three algorithms. Here, we set $U=0$, $\sigma^2 = 5$, $h^2 = 1$ and $\epsilon = 0.5$. The figure shows the convergence for three sparsity factors, namely $S = 0.4$, $S = 0.5$, and $S = 0.6$. As observed, the \rgtpK algorithm starts to track the (non-sparsified) distributed \ac{sgd} at $S=0.6$ while \tpK remains at a certain distance from the optimal solution.  
This behavior can be intuitively explained as follows: for both sparsification approaches, the aggregation of large local gradient entries gradually moves the initial point towards the global optimum. At a certain vicinity of the optimum, the impact of smaller gradient entries in convergence becomes more dominant. \tpK selects these entries only after large error accumulation, which due to learning rate scaling leads to oscillation around the optimum at a fixed distance. To avoid such oscillation, \tpK would need significant damping of the learning rate that can slow convergence. \rgtpK, however, selects the small (but dominant) gradient entries at lower error accumulation levels, which prevents large learning rate scaling. 

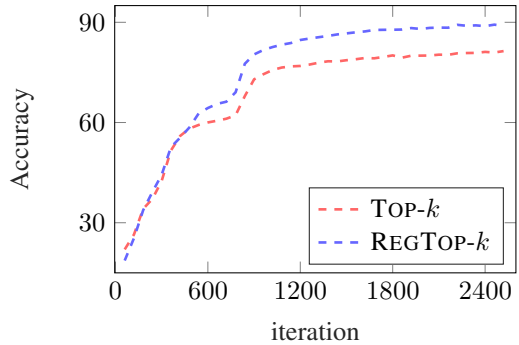
\begin{figure}
	\begin{center}
		\input{Figures/Accuracy_vs_Round.tex}
	\end{center}
	\vspace{-5mm}
	\caption{ResNet-18 on CIFAR-10 with $0.1\%$ sparsification.} 
	\vspace{-4mm}
	\label{fig:resnet}
\end{figure}
\subsection{Training ResNet-18 on CIFAR-10}
We now employ \rgtpK to train ResNet-18 on CIFAR-10, with data-points distributed evenly among $N=8$ workers. The clients compute their local gradients over mini-batches of size $20$. The aggregation is performed by arithmetic averaging. The learning rate is set to $\eta=0.01$, and 
the clients sparsify with $S=0.001$, i.e., $0.1\%$ sparsification. 

\paragraph{Numerical Results}
Figure~\ref{fig:resnet} shows the validation accuracy against the number of iterations for both \tpK and \rgtpK. To keep the comparison fair, we have considered the same initialization of the global model for both algorithms and identical batch samplers. 
As the figure shows, after the first 600 iterations, the model trained by \rgtpK sparsification starts to give strictly higher accuracy as compared with the one trained by standard \tpK. As the number of iterations surpasses 1500, the difference between the accuracy values exceeds $8\%$. 
This considerable gain indicates that \rgtpK can substantially improve the efficiency of sparsification in real-world applications.

\section{Conclusions}
Information collected during training can be used for efficient compression of local gradients in distributed learning. We have invoked this idea and developed a Bayesian framework to regularize the \tpK sparsification algorithm. Numerical investigations validate our derivations. \rgtpK can track the performance of non-sparsified distributed learning at significantly lower sparsity factors than \tpK. Interestingly, this gain is achieved with no considerable increase in computation complexity. This work can be extended in various respects. Most naturally, the proposed scheme can be extended to adaptive sparsification frameworks. 

\bibliographystyle{IEEEbib}
\bibliography{aaai24}

\end{document}

%% file: commands.tex
\usepackage{tikz,pgfplots}
\usepackage[nolist]{acronym}
\usepgfplotslibrary{fillbetween}

\newcommand{\setR}{\mathbb{R}}

\newcommand{\setS}{\mathbb{S}}

\newcommand{\setT}{\mathbb{T}}

\newcommand{\setD}{\mathbb{D}}
\newcommand{\setF}{\mathbb{F}}

\newcommand{\id}[2]{ _{ #2 \left[ #1\right]} }

\newcommand{\rmg}{\mathrm{g}}

\newcommand{\argmaxk}{\mathop{\mathrm{argmax}^k}}

\newcommand{\mal}{\mathcal{L}}

\newcommand{\bxx}{\mathbf{x}}

\newcommand{\bss}{\mathbf{s}}

\newcommand{\byy}{\mathbf{y}}

\newcommand{\bg}{{\mathbf{g}}}
\newcommand{\btheta}{\mathbf{w}}
\newcommand{\bdelta}{{\boldsymbol{\Delta}}}

\newcommand{\beps}{{\boldsymbol{\epsilon}}}

\newcommand{\bxi}{{\boldsymbol{\xi}}}

\newcommand{\bx}{{\boldsymbol{x}}}

\newcommand{\set}[1]{\left\lbrace#1\right\rbrace}

\newcommand{\brc}[1]{\left( #1 \right) }
\newcommand{\inner}[1]{\left\langle #1 \right\rangle }

\newcommand{\dbc}[1]{\left[ #1 \right] }

\newcommand{\bt}{{\mathbf{t}}}

\newcommand{\bzz}{{\mathbf{z}}}
\newcommand{\baa}{{\mathbf{a}}}

\newcommand{\dif}{\mathrm{d}}

\newcommand{\trp}{\mathsf{T}}

\newcommand{\Top}[2]{\mathrm{Top}_{#1} \left( #2 \right)}

\newcommand{\norm}[1]{\left\lVert #1 \right\rVert}

\newcommand{\abs}[1]{\lvert #1 \rvert}

\newtheorem{definition}{Definition}
\newtheorem{proposition}{Proposition}

%% file: Figures/Example_1.tex
%
%
\begin{tikzpicture}

\begin{axis}[%
	width=2.1in, 
	height=1.4in, 
	at={(1.262in,0.7in)},
	scale only axis,  
	xmin=-5,  
	xmax=105,  
	xtick={1, 50, 100},
	xticklabels={{$1$}, {$50$},  {$100$}},
	xlabel style={font=\color{white!15!black}},  
	xlabel={iteration},  
	ymode=log,  
	ymin=0.008,  
	ymax=0.4,  
	ytick={.1, .01, .001, .0001},
	yticklabels={{$10^{-1}$}, {$10^{-2}$}, {$10^{-3}$}, {$10^{-4}$}},
	ylabel style={font=\color{white!15!black}}, 
	ylabel={$F\brc{\btheta}$},
	yminorticks=true,
	axis background/.style={fill=white},
	legend style={at={(.97, .8)}, legend cell align=left, align=left, draw=white!15!black},
	]
\addplot[color=green!30!black, dashed, line width=1.0pt]
  table[row sep=crcr]{%
0    0.3132616875182228 \\
1    0.25370563548358493 \\
2    0.21191938042036657 \\
3    0.18129846918090053 \\
4    0.15804323758136132 \\
5    0.1398568080066663 \\
6    0.12528608965486968 \\
7    0.11337424038314028 \\
8    0.1034689093611886 \\
9    0.09511175245836298 \\
10    0.08797220867849892 \\
11    0.08180637527415294 \\
12    0.0764306505659697 \\
13    0.07170436026460299 \\
14    0.06751801481319772 \\
15    0.06378519268116806 \\
16    0.060436815801216404 \\
17    0.057417038108249906 \\
18    0.054680243640390785 \\
19    0.052188821732928624 \\
20    0.04991149548304802 \\
21    0.04782205010395581 \\
22    0.045898354330541784 \\
23    0.044121599333065656 \\
24    0.04247570097790551 \\
25    0.040946826103498146 \\
26    0.03952301390667582 \\
27    0.03819387096104483 \\
28    0.0369503237419168 \\
29    0.035784416433872646 \\
30    0.03468914467061875 \\
31    0.033658317994053624 \\
32    0.03268644542374661 \\
33    0.0317686397427149 \\
34    0.030900537032595432 \\
35    0.03007822870455887 \\
36    0.029298203824974715 \\
37    0.028557299966021972 \\
38    0.027852661150064607 \\
39    0.02718170172418573 \\
40    0.02654207521396218 \\
41    0.025931647375577092 \\
42    0.025348472801988146 \\
43    0.024790774549227044 \\
44    0.02425692633847888 \\
45    0.023745436962632 \\
46    0.02325493658581812 \\
47    0.022784164673679805 \\
48    0.022331959332745246 \\
49    0.02189724787100796 \\
50    0.021479038419856024 \\
51    0.021076412480933802 \\
52    0.02068851828117008 \\
53    0.02031456483572135 \\
54    0.019953816632528886 \\
55    0.01960558886398785 \\
56    0.019269243141239734 \\
57    0.01894418363513866 \\
58    0.018629853595216114 \\
59    0.018325732204210465 \\
60    0.018031331731074327 \\
61    0.017746194949979113 \\
62    0.01746989279681125 \\
63    0.017202022238085396 \\
64    0.016942204330177273 \\
65    0.01669008244936644 \\
66    0.016445320675427453 \\
67    0.016207602313469217 \\
68    0.015976628540442888 \\
69    0.01575211716423465 \\
70    0.01553380148458323 \\
71    0.015321429246215358 \\
72    0.015114761675619242 \\
73    0.014913572593770205 \\
74    0.014717647597919726 \\
75    0.01452678330626686 \\
76    0.01434078665994907 \\
77    0.014159474277347608 \\
78    0.013982671856197811 \\
79    0.01381021361942314 \\
80    0.013641941801021061 \\
81    0.013477706168659235 \\
82    0.013317363579966414 \\
83    0.013160777569781488 \\
84    0.013007817965867264 \\
85    0.012858360530830453 \\
86    0.012712286628187644 \\
87    0.01256948291070123 \\
88    0.012429841029272108 \\
89    0.01229325736082824 \\
90    0.012159632753777105 \\
91    0.012028872289717343 \\
92    0.011900885060210711 \\
93    0.01177558395751657 \\
94    0.011652885478281163 \\
95    0.01153270953925782 \\
96    0.011414979304205403 \\
97    0.011299621021184158 \\
98    0.01118656386952845 \\
99    0.011075739815829239 \\
100    0.01096708347831878 \\
};
\addlegendentry{No Sparsification}

\addplot[color=blue!60,   line width=1.0pt]
  table[row sep=crcr]{%
0    0.3132616875182228 \\
1    0.3132616875182228 \\
2    0.3132616875182228 \\
3    0.3132616875182228 \\
4    0.3132616875182228 \\
5    0.3132616875182228 \\
6    0.3132616875182228 \\
7    0.3132616875182228 \\
8    0.3132616875182228 \\
9    0.3132616875182228 \\
10    0.3132616875182228 \\
11    0.3132616875182228 \\
12    0.3132616875182228 \\
13    0.3132616875182228 \\
14    0.3132616875182228 \\
15    0.3132616875182228 \\
16    0.3132616875182228 \\
17    0.3132616875182228 \\
18    0.3132616875182228 \\
19    0.3132616875182228 \\
20    0.3132616875182228 \\
21    0.3132616875182228 \\
22    0.3132616875182228 \\
23    0.3132616875182228 \\
24    0.3132616875182228 \\
25    0.3132616875182228 \\
26    0.3132616875182228 \\
27    0.3132616875182228 \\
28    0.3132616875182228 \\
29    0.3132616875182228 \\
30    0.3132616875182228 \\
31    0.3132616875182228 \\
32    0.3132616875182228 \\
33    0.3132616875182228 \\
34    0.3132616875182228 \\
35    0.3132616875182228 \\
36    0.3132616875182228 \\
37    0.3132616875182228 \\
38    0.3132616875182228 \\
39    0.3132616875182228 \\
40    0.3132616875182228 \\
41    0.3132616875182228 \\
42    0.3132616875182228 \\
43    0.3132616875182228 \\
44    0.3132616875182228 \\
45    0.3132616875182228 \\
46    0.3132616875182228 \\
47    0.3132616875182228 \\
48    0.3132616875182228 \\
49    0.3132616875182228 \\
50    0.3132616875182228 \\
51    0.3132616875182228 \\
52    0.3132616875182228 \\
53    0.3132616875182228 \\
54    0.3132616875182228 \\
55    0.3132616875182228 \\
56    0.3132616875182228 \\
57    0.3132616875182228 \\
58    0.3132616875182228 \\
59    0.3132616875182228 \\
60    0.3132616875182228 \\
61    0.3132616875182228 \\
62    0.3132616875182228 \\
63    0.3132616875182228 \\
64    0.3132616875182228 \\
65    0.3132616875182228 \\
66    0.3132616875182228 \\
67    0.3132616875182228 \\
68    0.3132616875182228 \\
69    0.3132616875182228 \\
70    0.3132616875182228 \\
71    0.3132616875182228 \\
72    0.3132616875182228 \\
73    0.3132616875182228 \\
74    0.3132616875182228 \\
75    0.3132616875182228 \\
76    0.3132616875182228 \\
77    0.3132616875182228 \\
78    0.3132616875182228 \\
79    0.3132616875182228 \\
80    0.3132616875182228 \\
81    0.3132616875182228 \\
82    0.3132616875182228 \\
83    0.3132616875182228 \\
84    0.3132616875182228 \\
85    0.3132616875182228 \\
86    0.3132616875182228 \\
87    0.3132616875182228 \\
88    0.3132616875182228 \\
89    0.3132616875182228 \\
90    0.3132616875182228 \\
91    0.3132616875182228 \\
92    0.3132616875182228 \\
93    0.3132616875182228 \\
94    0.3132616875182228 \\
95    0.3132616875182228 \\
96    0.3132616875182228 \\
97    0.3132616875182228 \\
98    0.3132616875182228 \\
99    0.3132616875182228 \\
100    0.3132616875182228 \\
};
\addlegendentry{Top-$k$}

\addplot[dotted, color=orange,   line width=1.0pt]
table[row sep=crcr]{%
0    0.3132616875182228 \\
1    0.3132616875182228 \\
2    0.2043337658201848 \\
3    0.2043337658201848 \\
4    0.15061831680513285 \\
5    0.15061831680513285 \\
6    0.11901193829475559 \\
7    0.11901193829475559 \\
8    0.09827923815532132 \\
9    0.09827923815532132 \\
10    0.0836587674122629 \\
11    0.0836587674122629 \\
12    0.0728050227006296 \\
13    0.0728050227006296 \\
14    0.06443304245794443 \\
15    0.06443304245794443 \\
16    0.05778125146154424 \\
17    0.05778125146154424 \\
18    0.05237012303775674 \\
19    0.05237012303775674 \\
20    0.04788288969614484 \\
21    0.04788288969614484 \\
22    0.044101990592272655 \\
23    0.044101990592272655 \\
24    0.040873104154983735 \\
25    0.040873104154983735 \\
26    0.03808375384526146 \\
27    0.03808375384526146 \\
28    0.035650040413497196 \\
29    0.035650040413497196 \\
30    0.03350811680358467 \\
31    0.03350811680358467 \\
32    0.031608540574866076 \\
33    0.031608540574866076 \\
34    0.02991243184274419 \\
35    0.02991243184274419 \\
36    0.028388797814203216 \\
37    0.028388797814203216 \\
38    0.027012630811616627 \\
39    0.027012630811616627 \\
40    0.025763531034338132 \\
41    0.025763531034338132 \\
42    0.024624692672298228 \\
43    0.024624692672298228 \\
44    0.023582146298141097 \\
45    0.023582146298141097 \\
46    0.022624185052362626 \\
47    0.022624185052362626 \\
48    0.02174092464697344 \\
49    0.02174092464697344 \\
50    0.020923962156130235 \\
51    0.020923962156130235 \\
52    0.020166108661321534 \\
53    0.020166108661321534 \\
54    0.019461177757450203 \\
55    0.019461177757450203 \\
56    0.018803816766160122 \\
57    0.018803816766160122 \\
58    0.018189370926217766 \\
59    0.018189370926217766 \\
60    0.01761377328369794 \\
61    0.01761377328369794 \\
62    0.017073454783428418 \\
63    0.017073454783428418 \\
64    0.016565270367391437 \\
65    0.016565270367391437 \\
66    0.016086437852088745 \\
67    0.016086437852088745 \\
68    0.015634487079814494 \\
69    0.015634487079814494 \\
70    0.015207217384566514 \\
71    0.015207217384566514 \\
72    0.014802661828913502 \\
73    0.014802661828913502 \\
74    0.014419056987141193 \\
75    0.014419056987141193 \\
76    0.014054817296715297 \\
77    0.014054817296715297 \\
78    0.013708513192294358 \\
79    0.013708513192294358 \\
80    0.013378852387246145 \\
81    0.013378852387246145 \\
82    0.013064663786598388 \\
83    0.013064663786598388 \\
84    0.01276488360981402 \\
85    0.01276488360981402 \\
86    0.012478543377248165 \\
87    0.012478543377248165 \\
88    0.012204759474737894 \\
89    0.012204759474737894 \\
90    0.011942724059695136 \\
91    0.011942724059695136 \\
92    0.011691697111769409 \\
93    0.011691697111769409 \\
94    0.011450999463496047 \\
95    0.011450999463496047 \\
96    0.011220006672851227 \\
97    0.011220006672851227 \\
98    0.01099814362142096 \\
99    0.01099814362142096 \\
100    0.010784879739895448 \\
};
\addlegendentry{RegTop-$k$}

\end{axis}
\end{tikzpicture}%

%% file: Figures/Accuracy_vs_Round.tex
%
%
\begin{tikzpicture}

\begin{axis}[%
	width=2.1in, 
	height=1.4in, 
	at={(1.262in,0.7in)},
	scale only axis,  
	xmin=-5,  
	xmax=2600,  
	xtick={0, 600,  1200,  1800,  2400},
	xticklabels={{$0$}, {$600$}, {$1200$},  {$1800$},  {$2400$}},
	xlabel style={font=\color{white!15!black}},  
	xlabel={iteration},    
	ymin=15,  
	ymax=95,  
	ytick={30, 60, 90},
	yticklabels={{$30$}, {$60$}, {$90$}},
	ylabel style={font=\color{white!15!black}}, 
	ylabel={Accuracy},
	yminorticks=true,
	axis background/.style={fill=white},
	legend style={at={(.97, .32)}, legend cell align=left, align=left, draw=white!15!black},
	]
\addplot[color=red!60, dashed, line width=1.0pt]
  table[row sep=crcr]{%
60	22.009994506835938 \\
120	26.460000610351564 \\
180	34.15000915527344 \\
240	37.150003051757814 \\
300	42.499996948242185 \\
360	51.61998291015625 \\
420	55.9800048828125 \\
480	58.18998413085937 \\
540	59.27003173828125 \\
600	60.020074462890626 \\
660	60.53999633789063 \\
720	61.139984130859375 \\
780	62.160003662109375 \\
840	67.92001953125 \\
900	72.92001342773438 \\
960	74.43994750976563 \\
1020	75.72000122070312 \\
1080	76.45997924804688 \\
1140	76.80001831054688 \\
1200	76.92996826171876 \\
1260	77.34998168945313 \\
1320	77.90997314453125 \\
1380	78.33004760742188 \\
1440	78.28001098632812 \\
1500	78.64998168945313 \\
1560	78.93995971679688 \\
1620	79.30997314453126 \\
1680	79.2199951171875 \\
1740	79.62999877929687 \\
1800	80.09000854492187 \\
1860	79.39995727539062 \\
1920	80.000048828125 \\
1980	80.02002563476563 \\
2040	80.05001220703124 \\
2100	80.3099853515625 \\
2160	80.2599609375 \\
2220	80.7199951171875 \\
2280	80.79998779296875 \\
2340	80.83998413085938 \\
2400	81.1400146484375 \\
2460	80.95000610351562 \\
2520	81.3800048828125 \\
};
\addlegendentry{\tpK}

\addplot[color=blue!60, dashed,   line width=1.0pt]
  table[row sep=crcr]{%
60	18.689990234375 \\
120	25.37000274658203 \\
180	33.76001892089844 \\
240	39.20997009277344 \\
300	44.34001159667969 \\
360	52.6199951171875 \\
420	55.47001953125 \\
480	58.519964599609374 \\
540	62.69999389648437 \\
600	64.33997802734375 \\
660	65.5699951171875 \\
720	66.23001708984376 \\
780	69.0500244140625 \\
840	77.58997802734375 \\
900	80.38004150390626 \\
960	81.69994506835937 \\
1020	82.59002075195312 \\
1080	83.30000610351563 \\
1140	83.7999755859375 \\
1200	84.70003051757813 \\
1260	85.1400146484375 \\
1320	85.539990234375 \\
1380	85.90997924804688 \\
1440	86.37000732421875 \\
1500	86.6800048828125 \\
1560	86.99998168945312 \\
1620	87.27000122070312 \\
1680	87.70003051757813 \\
1740	87.74996337890624 \\
1800	87.77003173828125 \\
1860	87.82998657226562 \\
1920	88.27001342773437 \\
1980	88.01000366210937 \\
2040	88.31998901367187 \\
2100	88.41007080078126 \\
2160	88.37999267578125 \\
2220	89.24005737304688 \\
2280	88.95003662109374 \\
2340	89.08003540039063 \\
2400	88.8500244140625 \\
2460	89.2900390625 \\
2520	88.92003784179687 \\
};
\addlegendentry{\rgtpK}

\end{axis}
\end{tikzpicture}%